\documentclass{article}

\usepackage[preprint]{corl_2023}

\usepackage[utf8]{inputenc} 
\usepackage[T1]{fontenc}    
\usepackage{hyperref}       
\usepackage{url}            
\usepackage{booktabs}       
\usepackage{amsfonts}       
\usepackage{amsmath}
\usepackage{nicefrac}       
\usepackage{microtype}      
\usepackage[dvipsnames]{xcolor}         
\usepackage[capitalise]{cleveref}
\usepackage{bm}
\usepackage{graphicx}
\usepackage[ruled,vlined, linesnumbered, noend]{algorithm2e}
\usepackage{subcaption}
\usepackage[moderate]{savetrees}
\usepackage{soul}           

\DeclareMathAlphabet{\mathsfit}{\encodingdefault}{\sfdefault}{m}{sl}
\SetMathAlphabet{\mathsfit}{bold}{\encodingdefault}{\sfdefault}{bx}{n}

\hypersetup{
    colorlinks=true,
    linkcolor = blue,
    anchorcolor = blue,
    citecolor = blue,
    filecolor = blue,
    urlcolor = blue
}

\title{Value function estimation using conditional diffusion models for control}

\author{%
  Bogdan Mazoure\\
  Apple\\
  \And
  Walter Talbott \\
  Apple \\
  \And
  Miguel Angel Bautista \\
  Apple \\
  \And
  Devon Hjelm \\
  Apple \\
  \And
  Alexander Toshev \\
  Apple
  \And
  Josh Susskind \\
  Apple \\
}

\newcommand\longmethod{Diffused Value Function}
\newcommand\shortmethod{DVF}

\begin{document}

\newcommand{\figleft}{{\em (Left)}}
\newcommand{\figcenter}{{\em (Center)}}
\newcommand{\figright}{{\em (Right)}}
\newcommand{\figtop}{{\em (Top)}}
\newcommand{\figbottom}{{\em (Bottom)}}
\newcommand{\captiona}{{\em (a)}}
\newcommand{\captionb}{{\em (b)}}
\newcommand{\captionc}{{\em (c)}}
\newcommand{\captiond}{{\em (d)}}

\newcommand{\newterm}[1]{{\bf #1}}

\newtheorem{obs}{Observation}
\newtheorem{prop}{Proposition}

\def\figref#1{figure~\ref{#1}}
\def\Figref#1{Figure~\ref{#1}}
\def\twofigref#1#2{figures \ref{#1} and \ref{#2}}
\def\quadfigref#1#2#3#4{figures \ref{#1}, \ref{#2}, \ref{#3} and \ref{#4}}
\def\secref#1{section~\ref{#1}}
\def\Secref#1{Section~\ref{#1}}
\def\twosecrefs#1#2{sections \ref{#1} and \ref{#2}}
\def\secrefs#1#2#3{sections \ref{#1}, \ref{#2} and \ref{#3}}
\def\eqref#1{equation~\ref{#1}}
\def\Eqref#1{Equation~\ref{#1}}
\def\plaineqref#1{\ref{#1}}
\def\chapref#1{chapter~\ref{#1}}
\def\Chapref#1{Chapter~\ref{#1}}
\def\rangechapref#1#2{chapters\ref{#1}--\ref{#2}}
\def\algref#1{algorithm~\ref{#1}}
\def\Algref#1{Algorithm~\ref{#1}}
\def\twoalgref#1#2{algorithms \ref{#1} and \ref{#2}}
\def\Twoalgref#1#2{Algorithms \ref{#1} and \ref{#2}}
\def\partref#1{part~\ref{#1}}
\def\Partref#1{Part~\ref{#1}}
\def\twopartref#1#2{parts \ref{#1} and \ref{#2}}

\def\ceil#1{\lceil #1 \rceil}
\def\floor#1{\lfloor #1 \rfloor}
\def\1{\bm{1}}
\newcommand{\train}{\mathcal{D}}
\newcommand{\valid}{\mathcal{D_{\mathrm{valid}}}}
\newcommand{\test}{\mathcal{D_{\mathrm{test}}}}

\def\eps{{\epsilon}}

\def\reta{{\textnormal{$\eta$}}}
\def\ra{{\textnormal{a}}}
\def\rb{{\textnormal{b}}}
\def\rc{{\textnormal{c}}}
\def\rd{{\textnormal{d}}}
\def\re{{\textnormal{e}}}
\def\rf{{\textnormal{f}}}
\def\rg{{\textnormal{g}}}
\def\rh{{\textnormal{h}}}
\def\ri{{\textnormal{i}}}
\def\rj{{\textnormal{j}}}
\def\rk{{\textnormal{k}}}
\def\rl{{\textnormal{l}}}
\def\rn{{\textnormal{n}}}
\def\ro{{\textnormal{o}}}
\def\rp{{\textnormal{p}}}
\def\rq{{\textnormal{q}}}
\def\rr{{\textnormal{r}}}
\def\rs{{\textnormal{s}}}
\def\rt{{\textnormal{t}}}
\def\ru{{\textnormal{u}}}
\def\rv{{\textnormal{v}}}
\def\rw{{\textnormal{w}}}
\def\rx{{\textnormal{x}}}
\def\ry{{\textnormal{y}}}
\def\rz{{\textnormal{z}}}

\def\rvepsilon{{\mathbf{\epsilon}}}
\def\rvtheta{{\mathbf{\theta}}}
\def\rva{{\mathbf{a}}}
\def\rvb{{\mathbf{b}}}
\def\rvc{{\mathbf{c}}}
\def\rvd{{\mathbf{d}}}
\def\rve{{\mathbf{e}}}
\def\rvf{{\mathbf{f}}}
\def\rvg{{\mathbf{g}}}
\def\rvh{{\mathbf{h}}}
\def\rvu{{\mathbf{i}}}
\def\rvj{{\mathbf{j}}}
\def\rvk{{\mathbf{k}}}
\def\rvl{{\mathbf{l}}}
\def\rvm{{\mathbf{m}}}
\def\rvn{{\mathbf{n}}}
\def\rvo{{\mathbf{o}}}
\def\rvp{{\mathbf{p}}}
\def\rvq{{\mathbf{q}}}
\def\rvr{{\mathbf{r}}}
\def\rvs{{\mathbf{s}}}
\def\rvt{{\mathbf{t}}}
\def\rvu{{\mathbf{u}}}
\def\rvv{{\mathbf{v}}}
\def\rvw{{\mathbf{w}}}
\def\rvx{{\mathbf{x}}}
\def\rvy{{\mathbf{y}}}
\def\rvz{{\mathbf{z}}}

\def\erva{{\textnormal{a}}}
\def\ervb{{\textnormal{b}}}
\def\ervc{{\textnormal{c}}}
\def\ervd{{\textnormal{d}}}
\def\erve{{\textnormal{e}}}
\def\ervf{{\textnormal{f}}}
\def\ervg{{\textnormal{g}}}
\def\ervh{{\textnormal{h}}}
\def\ervi{{\textnormal{i}}}
\def\ervj{{\textnormal{j}}}
\def\ervk{{\textnormal{k}}}
\def\ervl{{\textnormal{l}}}
\def\ervm{{\textnormal{m}}}
\def\ervn{{\textnormal{n}}}
\def\ervo{{\textnormal{o}}}
\def\ervp{{\textnormal{p}}}
\def\ervq{{\textnormal{q}}}
\def\ervr{{\textnormal{r}}}
\def\ervs{{\textnormal{s}}}
\def\ervt{{\textnormal{t}}}
\def\ervu{{\textnormal{u}}}
\def\ervv{{\textnormal{v}}}
\def\ervw{{\textnormal{w}}}
\def\ervx{{\textnormal{x}}}
\def\ervy{{\textnormal{y}}}
\def\ervz{{\textnormal{z}}}

\def\rmA{{\mathbf{A}}}
\def\rmB{{\mathbf{B}}}
\def\rmC{{\mathbf{C}}}
\def\rmD{{\mathbf{D}}}
\def\rmE{{\mathbf{E}}}
\def\rmF{{\mathbf{F}}}
\def\rmG{{\mathbf{G}}}
\def\rmH{{\mathbf{H}}}
\def\rmI{{\mathbf{I}}}
\def\rmJ{{\mathbf{J}}}
\def\rmK{{\mathbf{K}}}
\def\rmL{{\mathbf{L}}}
\def\rmM{{\mathbf{M}}}
\def\rmN{{\mathbf{N}}}
\def\rmO{{\mathbf{O}}}
\def\rmP{{\mathbf{P}}}
\def\rmQ{{\mathbf{Q}}}
\def\rmR{{\mathbf{R}}}
\def\rmS{{\mathbf{S}}}
\def\rmT{{\mathbf{T}}}
\def\rmU{{\mathbf{U}}}
\def\rmV{{\mathbf{V}}}
\def\rmW{{\mathbf{W}}}
\def\rmX{{\mathbf{X}}}
\def\rmY{{\mathbf{Y}}}
\def\rmZ{{\mathbf{Z}}}

\def\ermA{{\textnormal{A}}}
\def\ermB{{\textnormal{B}}}
\def\ermC{{\textnormal{C}}}
\def\ermD{{\textnormal{D}}}
\def\ermE{{\textnormal{E}}}
\def\ermF{{\textnormal{F}}}
\def\ermG{{\textnormal{G}}}
\def\ermH{{\textnormal{H}}}
\def\ermI{{\textnormal{I}}}
\def\ermJ{{\textnormal{J}}}
\def\ermK{{\textnormal{K}}}
\def\ermL{{\textnormal{L}}}
\def\ermM{{\textnormal{M}}}
\def\ermN{{\textnormal{N}}}
\def\ermO{{\textnormal{O}}}
\def\ermP{{\textnormal{P}}}
\def\ermQ{{\textnormal{Q}}}
\def\ermR{{\textnormal{R}}}
\def\ermS{{\textnormal{S}}}
\def\ermT{{\textnormal{T}}}
\def\ermU{{\textnormal{U}}}
\def\ermV{{\textnormal{V}}}
\def\ermW{{\textnormal{W}}}
\def\ermX{{\textnormal{X}}}
\def\ermY{{\textnormal{Y}}}
\def\ermZ{{\textnormal{Z}}}

\def\vzero{{\bm{0}}}
\def\vone{{\bm{1}}}
\def\vmu{{\bm{\mu}}}
\def\vtheta{{\bm{\theta}}}
\def\va{{\bm{a}}}
\def\vb{{\bm{b}}}
\def\vc{{\bm{c}}}
\def\vd{{\bm{d}}}
\def\ve{{\bm{e}}}
\def\vf{{\bm{f}}}
\def\vg{{\bm{g}}}
\def\vh{{\bm{h}}}
\def\vi{{\bm{i}}}
\def\vj{{\bm{j}}}
\def\vk{{\bm{k}}}
\def\vl{{\bm{l}}}
\def\vm{{\bm{m}}}
\def\vn{{\bm{n}}}
\def\vo{{\bm{o}}}
\def\vp{{\bm{p}}}
\def\vq{{\bm{q}}}
\def\vr{{\bm{r}}}
\def\vs{{\bm{s}}}
\def\vt{{\bm{t}}}
\def\vu{{\bm{u}}}
\def\vv{{\bm{v}}}
\def\vw{{\bm{w}}}
\def\vx{{\bm{x}}}
\def\vy{{\bm{y}}}
\def\vz{{\bm{z}}}

\def\evalpha{{\alpha}}
\def\evbeta{{\beta}}
\def\evepsilon{{\epsilon}}
\def\evlambda{{\lambda}}
\def\evomega{{\omega}}
\def\evmu{{\mu}}
\def\evpsi{{\psi}}
\def\evsigma{{\sigma}}
\def\evtheta{{\theta}}
\def\eva{{a}}
\def\evb{{b}}
\def\evc{{c}}
\def\evd{{d}}
\def\eve{{e}}
\def\evf{{f}}
\def\evg{{g}}
\def\evh{{h}}
\def\evi{{i}}
\def\evj{{j}}
\def\evk{{k}}
\def\evl{{l}}
\def\evm{{m}}
\def\evn{{n}}
\def\evo{{o}}
\def\evp{{p}}
\def\evq{{q}}
\def\evr{{r}}
\def\evs{{s}}
\def\evt{{t}}
\def\evu{{u}}
\def\evv{{v}}
\def\evw{{w}}
\def\evx{{x}}
\def\evy{{y}}
\def\evz{{z}}

\def\mA{{\bm{A}}}
\def\mB{{\bm{B}}}
\def\mC{{\bm{C}}}
\def\mD{{\bm{D}}}
\def\mE{{\bm{E}}}
\def\mF{{\bm{F}}}
\def\mG{{\bm{G}}}
\def\mH{{\bm{H}}}
\def\mI{{\bm{I}}}
\def\mJ{{\bm{J}}}
\def\mK{{\bm{K}}}
\def\mL{{\bm{L}}}
\def\mM{{\bm{M}}}
\def\mN{{\bm{N}}}
\def\mO{{\bm{O}}}
\def\mP{{\bm{P}}}
\def\mQ{{\bm{Q}}}
\def\mR{{\bm{R}}}
\def\mS{{\bm{S}}}
\def\mT{{\bm{T}}}
\def\mU{{\bm{U}}}
\def\mV{{\bm{V}}}
\def\mW{{\bm{W}}}
\def\mX{{\bm{X}}}
\def\mY{{\bm{Y}}}
\def\mZ{{\bm{Z}}}
\def\mBeta{{\bm{\beta}}}
\def\mPhi{{\bm{\Phi}}}
\def\mLambda{{\bm{\Lambda}}}
\def\mSigma{{\bm{\Sigma}}}

\newcommand{\tens}[1]{\bm{\mathsfit{#1}}}
\def\tA{{\tens{A}}}
\def\tB{{\tens{B}}}
\def\tC{{\tens{C}}}
\def\tD{{\tens{D}}}
\def\tE{{\tens{E}}}
\def\tF{{\tens{F}}}
\def\tG{{\tens{G}}}
\def\tH{{\tens{H}}}
\def\tI{{\tens{I}}}
\def\tJ{{\tens{J}}}
\def\tK{{\tens{K}}}
\def\tL{{\tens{L}}}
\def\tM{{\tens{M}}}
\def\tN{{\tens{N}}}
\def\tO{{\tens{O}}}
\def\tP{{\tens{P}}}
\def\tQ{{\tens{Q}}}
\def\tR{{\tens{R}}}
\def\tS{{\tens{S}}}
\def\tT{{\tens{T}}}
\def\tU{{\tens{U}}}
\def\tV{{\tens{V}}}
\def\tW{{\tens{W}}}
\def\tX{{\tens{X}}}
\def\tY{{\tens{Y}}}
\def\tZ{{\tens{Z}}}

\def\gA{{\mathcal{A}}}
\def\gB{{\mathcal{B}}}
\def\gC{{\mathcal{C}}}
\def\gD{{\mathcal{D}}}
\def\gE{{\mathcal{E}}}
\def\gF{{\mathcal{F}}}
\def\gG{{\mathcal{G}}}
\def\gH{{\mathcal{H}}}
\def\gI{{\mathcal{I}}}
\def\gJ{{\mathcal{J}}}
\def\gK{{\mathcal{K}}}
\def\gL{{\mathcal{L}}}
\def\gM{{\mathcal{M}}}
\def\gN{{\mathcal{N}}}
\def\gO{{\mathcal{O}}}
\def\gP{{\mathcal{P}}}
\def\gQ{{\mathcal{Q}}}
\def\gR{{\mathcal{R}}}
\def\gS{{\mathcal{S}}}
\def\gT{{\mathcal{T}}}
\def\gU{{\mathcal{U}}}
\def\gV{{\mathcal{V}}}
\def\gW{{\mathcal{W}}}
\def\gX{{\mathcal{X}}}
\def\gY{{\mathcal{Y}}}
\def\gZ{{\mathcal{Z}}}

\def\sA{{\mathbb{A}}}
\def\sB{{\mathbb{B}}}
\def\sC{{\mathbb{C}}}
\def\sD{{\mathbb{D}}}
\def\sF{{\mathbb{F}}}
\def\sG{{\mathbb{G}}}
\def\sH{{\mathbb{H}}}
\def\sI{{\mathbb{I}}}
\def\sJ{{\mathbb{J}}}
\def\sK{{\mathbb{K}}}
\def\sL{{\mathbb{L}}}
\def\sM{{\mathbb{M}}}
\def\sN{{\mathbb{N}}}
\def\sO{{\mathbb{O}}}
\def\sP{{\mathbb{P}}}
\def\sQ{{\mathbb{Q}}}
\def\sR{{\mathbb{R}}}
\def\sS{{\mathbb{S}}}
\def\sT{{\mathbb{T}}}
\def\sU{{\mathbb{U}}}
\def\sV{{\mathbb{V}}}
\def\sW{{\mathbb{W}}}
\def\sX{{\mathbb{X}}}
\def\sY{{\mathbb{Y}}}
\def\sZ{{\mathbb{Z}}}

\def\emLambda{{\Lambda}}
\def\emA{{A}}
\def\emB{{B}}
\def\emC{{C}}
\def\emD{{D}}
\def\emE{{E}}
\def\emF{{F}}
\def\emG{{G}}
\def\emH{{H}}
\def\emI{{I}}
\def\emJ{{J}}
\def\emK{{K}}
\def\emL{{L}}
\def\emM{{M}}
\def\emN{{N}}
\def\emO{{O}}
\def\emP{{P}}
\def\emQ{{Q}}
\def\emR{{R}}
\def\emS{{S}}
\def\emT{{T}}
\def\emU{{U}}
\def\emV{{V}}
\def\emW{{W}}
\def\emX{{X}}
\def\emY{{Y}}
\def\emZ{{Z}}
\def\emSigma{{\Sigma}}

\newcommand{\etens}[1]{\mathsfit{#1}}
\def\etLambda{{\etens{\Lambda}}}
\def\etA{{\etens{A}}}
\def\etB{{\etens{B}}}
\def\etC{{\etens{C}}}
\def\etD{{\etens{D}}}
\def\etE{{\etens{E}}}
\def\etF{{\etens{F}}}
\def\etG{{\etens{G}}}
\def\etH{{\etens{H}}}
\def\etI{{\etens{I}}}
\def\etJ{{\etens{J}}}
\def\etK{{\etens{K}}}
\def\etL{{\etens{L}}}
\def\etM{{\etens{M}}}
\def\etN{{\etens{N}}}
\def\etO{{\etens{O}}}
\def\etP{{\etens{P}}}
\def\etQ{{\etens{Q}}}
\def\etR{{\etens{R}}}
\def\etS{{\etens{S}}}
\def\etT{{\etens{T}}}
\def\etU{{\etens{U}}}
\def\etV{{\etens{V}}}
\def\etW{{\etens{W}}}
\def\etX{{\etens{X}}}
\def\etY{{\etens{Y}}}
\def\etZ{{\etens{Z}}}

\newcommand{\pdata}{p_{\rm{data}}}
\newcommand{\ptrain}{\hat{p}_{\rm{data}}}
\newcommand{\Ptrain}{\hat{P}_{\rm{data}}}
\newcommand{\pmodel}{p_{\rm{model}}}
\newcommand{\Pmodel}{P_{\rm{model}}}
\newcommand{\ptildemodel}{\tilde{p}_{\rm{model}}}
\newcommand{\pencode}{p_{\rm{encoder}}}
\newcommand{\pdecode}{p_{\rm{decoder}}}
\newcommand{\precons}{p_{\rm{reconstruct}}}

\newcommand{\laplace}{\mathrm{Laplace}} 

\newcommand{\E}{\mathbb{E}}
\newcommand{\Ls}{\mathcal{L}}
\newcommand{\R}{\mathbb{R}}
\newcommand{\emp}{\tilde{p}}
\newcommand{\lr}{\alpha}
\newcommand{\reg}{\lambda}
\newcommand{\rect}{\mathrm{rectifier}}
\newcommand{\softmax}{\mathrm{softmax}}
\newcommand{\sigmoid}{\sigma}
\newcommand{\softplus}{\zeta}
\newcommand{\KL}{D_{\mathrm{KL}}}
\newcommand{\Var}{\mathrm{Var}}
\newcommand{\standarderror}{\mathrm{SE}}
\newcommand{\Cov}{\mathrm{Cov}}
\newcommand{\normlzero}{L^0}
\newcommand{\normlone}{L^1}
\newcommand{\normltwo}{L^2}
\newcommand{\normlp}{L^p}
\newcommand{\normmax}{L^\infty}

\newcommand{\parents}{Pa} 

\let\ab\allowbreak

\newcommand{\cA}{\mathcal{A}}
\newcommand{\cB}{\mathcal{B}}
\newcommand{\cC}{\mathcal{C}}
\newcommand{\cD}{\mathcal{D}}
\newcommand{\cE}{\mathcal{E}}
\newcommand{\cF}{\mathcal{F}}
\newcommand{\cG}{\mathcal{G}}
\newcommand{\cH}{\mathcal{H}}
\newcommand{\cI}{\mathcal{I}}
\newcommand{\cJ}{\mathcal{J}}
\newcommand{\cK}{\mathcal{K}}
\newcommand{\cL}{\mathcal{L}}
\newcommand{\cM}{\mathcal{M}}
\newcommand{\cN}{\mathcal{N}}
\newcommand{\cO}{\mathcal{O}}
\newcommand{\cP}{\mathcal{P}}
\newcommand{\cQ}{\mathcal{Q}}
\newcommand{\cR}{\mathcal{R}}
\newcommand{\cS}{\mathcal{S}}
\newcommand{\cT}{\mathcal{T}}
\newcommand{\cU}{\mathcal{U}}
\newcommand{\cV}{\mathcal{V}}
\newcommand{\cW}{\mathcal{W}}
\newcommand{\cX}{\mathcal{X}}
\newcommand{\cY}{\mathcal{Y}}
\newcommand{\cZ}{\mathcal{Z}}

\newcommand{\Real}{\mathbb{R}}
\newcommand{\Nat}{\mathbb{N}}

\newtheorem{theorem}{Theorem}
\newtheorem{corollary}{Corollary}%
\newtheorem{lemma}{Lemma}
\newtheorem{example}{Example}
\newtheorem{definition}{Definition}
\newtheorem{assumption}{Assumption}
\newtheorem{proof}{Proof}

\newtheorem{remark}{Remark}

%
\renewcommand{\vec}[1]{\ensuremath{\bm{#1}}}
\newcommand{\vecs}[1]{\ensuremath{\mathbf{\boldsymbol{#1}}}}
\newcommand{\mat}[1]{\ensuremath{\mathbf{#1}}}
\newcommand{\mats}[1]{\ensuremath{\mathbf{\boldsymbol{#1}}}}
\newcommand{\ten}[1]{\mat{\ensuremath{\boldsymbol{\mathcal{#1}}}}}
\maketitle

\begin{abstract}
A fairly reliable trend in deep reinforcement learning is that the performance scales with the number of parameters, provided a complimentary scaling in amount of training data. As the appetite for large models increases, it is imperative to address, sooner than later, the potential problem of running out of high-quality demonstrations.
In this case, instead of collecting only new data via costly human demonstrations or risking a simulation-to-real transfer with uncertain effects, it would be beneficial to leverage vast amounts of readily-available low-quality data. 
Since classical control algorithms such as behavior cloning or temporal difference learning cannot be used on reward-free or action-free data out-of-the-box, this solution warrants novel training paradigms for continuous control. 
We propose a simple algorithm called \longmethod{} (\shortmethod), which learns a joint multi-step model of the environment-robot interaction dynamics using a diffusion model. This model can be efficiently learned from state sequences (i.e., without access to reward functions nor actions), and subsequently used to estimate the value of each action out-of-the-box. We show how \shortmethod{} can be used to efficiently capture the state visitation measure for multiple controllers, and show promising qualitative and quantitative results on challenging robotics benchmarks.
\end{abstract}

\section{Introduction}
The success of foundation models~\citep{chowdhery2022palm, touvron2023llama} is often attributed to their size~\citep{kaplan2020scaling} and abundant training data, a handful of which is usually annotated by a preference model trained on human feedback~\citep{ouyang2022training}. Similarly, the robotics community has seen a surge in large multimodal learners~\citep{brohan2023can,stone2023open,driess2023palm}, which also require vast amounts of high-quality training demonstrations. What can we do when annotating demonstrations is prohibitively costly, and the sim2real gap is too large? Recent works show that partially pre-training the controller on large amounts of low-returns data with missing information can help accelerate learning from optimal demonstration~\citep{baker2022video,fan2022minedojo}. 
A major drawback of these works lies in the compounding prediction error: training a preference model on optimal demonstrations and subsequently using this model in reinforcement learning (RL) or behavior cloning (BC) approaches includes both the uncertainty from the preference bootstrapping, as well as the RL algorithm itself. Instead, we opt for a different path: decompose the value function, a fundamental quantity for continuous control, into components that depend only on states, only on rewards, and only on actions. These individual pieces can then be trained separately on different subsets of available data, and re-combined together to construct a value function estimate, as shown in later sections.

Factorizing the value function into dynamics, decision and reward components poses a major challenge, since it requires disentangling the non-stationarity induced by the controller from that of the dynamical system. Model-based approaches address this problem by learning a differentiable transition model of the dynamic system, through which the information from the controller can be propagated~\citep{yu2020mopo, argenson2020model, kidambi2020morel, yu2021combo}. While these approaches can work well on some benchmarks, they can be complex and expensive: the model must predict high-dimensional observations, and determining the value of an action may require unrolling the model for multiple steps into the future.

In this paper, we show how we can estimate the environment dynamics in an efficient way while avoiding the dependence of model-based rollouts on the episode horizon. The model learned by our method \emph{(1)} does not require predicting high-dimensional observations at every timestep, \emph{(2)} directly predicts the future state without the need of autoregressive unrolls and \emph{(3)} can be used to estimate the value function without requiring expensive rollouts or temporal difference learning, nor does it need action or reward labels during the pre-training phase. Precisely, we learn a generative model of the discounted state occupancy measure, i.e. a function which takes in a state, action and timestep, and returns a future state proportional to the likelihood of visiting that future state under some fixed policy. This occupancy measure has a resemblance to successor features~\citep{dayan1993improving}, and can be seen as its generative, normalized version. By scoring these future states by the corresponding rewards, we form an unbiased estimate of the value function.  We name our proposed algorithm \longmethod{} (\shortmethod). Because~\shortmethod{} represents multi-step transitions implicitly, it avoids having to predict high-dimensional observations at every timestep and thus scales to long-horizon tasks with high-dimensional observations. Using the same algorithm, we can handle settings where reward-free and action-free data is provided, which cannot be directly handled by classical TD-based methods. Specifically, the generative model can be pre-trained on sequences of states without the need for reward or action labels, provided that some representation of the data generating process (i.e., logging policy) is known.

We highlight the strengths of~\shortmethod{} both qualitatively and quantitatively on challenging robotic tasks, and show how generative models can be used to accelerate \emph{tabula rasa} learning.

\section{Preliminaries}

\paragraph{Reinforcement learning}
Let $M$ be a Markov decision process (MDP) defined by the tuple $M=\langle \cS,S_0,\cA,\cT, r, \gamma \rangle$, where $\cS$ is a state space, $S_0\subseteq \cS$ is the set of starting states, $\cA$ is an action space, $\cT=p(\cdot|s_t,a_t):\cS\times \cA \to \Delta(\cS)$ is a one-step transition function\footnote{$\Delta(\cX)$ denotes the entire set of distributions over the space $\cX$.}, $r:\cS\times \cA \to [r_\text{min},r_\text{max}]$ is a reward function and $\gamma\in [0,1)$ is a discount factor. The system starts in one of the initial states $s_0\in S_0$. At every timestep $t>0$, the policy $\pi:\cS \to \Delta(\cA)$ samples an action $a_t\sim \pi(\cdot|s_t)$. The environment transitions into a next state $s_{t+1} \sim \cT(\cdot|s_t,a_t)$ and emits a reward $r_t=r(s_t,a_t)$. The aim is to learn a Markovian policy $\pi(a \mid s)$ that maximizes the return, defined as discounted sum of rewards, over an episode of length $H$:
\begin{equation}
    \max_{\pi\in \Pi} \E_{p^\pi_{0:H},S_0}\left[\sum_{t=0}^H \gamma^t r(s_t, a_t) \right],
    \label{eq:rl_maximize_rewards}
\end{equation}
where $p^\pi_{t:t+K}$ denotes the joint distribution of $\{s_{t+k},a_{t+k}\}_{k=1}^K$ obtained by rolling-out $\pi$ in the environment for $K$ timesteps starting at timestep $t$. To solve~\cref{eq:rl_maximize_rewards}, value-based RL algorithms estimate the future expected discounted sum of rewards, known as the \emph{value function}:
\begin{equation}
    Q^\pi(s_t,a_t)=\mathbb{E}_{p_t^\pi}\left[\sum_{k=1}^{H}\gamma^{k-1}r(s_{t+k},a_{t+k})|s_t,a_t \right],
    \label{eq:q_value}
\end{equation}
for $s_t\in \cS,a_t\in \cA$, and $V^\pi(s_t)=\mathbb{E}_\pi[Q(s_t,a_t)]$. Alternatively, the value function can be written as the expectation of the reward over the discounted occupancy measure:
\begin{align}
    Q^\pi(s_t,a_t)=\frac{1-\gamma^{H-t}}{1-\gamma}\mathbb{E}_{s,a\sim \rho^\pi(s_t,a_t),\pi(s)}[r(s,a)]
     \label{eq:occupancy_q_value}
\end{align}
where 
$\rho^\pi(s|s_t,a_t)=(1-\gamma)\sum_{\Delta t=1}^H\gamma^{\Delta t -1}\rho^\pi(s|s_t,a_t,\Delta t,\pi)$ and $\rho^\pi(s|s_t,a_t,\Delta t,\pi)=\mathbb{P}[S_{t+\Delta t}=s|s_t,a_t;\pi]$ as defined in~\cite{janner2020generative}.

This decomposition of the value function has been shown to be useful in previous works based on the successor representation~\citep{dayan1993improving,barreto2016successor} and $\gamma$-models~\citep{janner2020generative}, and we will leverage this formulation to build a diffusion-based estimate of the value function below.

\paragraph{Diffusion models}
Diffusion models form a class of latent variable models~\citep{sohl2015deep} which represent the distribution of the data as an iterative process: 
\begin{align}
\vec{x}_0\sim p(\vec{x}_0)=\mathbb{E}_{p_{\theta}(\vec{x}_{1:T})}[p_{\theta}(\vec{x}_0|\vec{x}_{1:T})]=p(\vec{x}_T)\prod_{t_d=1}^Tp_\theta(\vec{x}_{t_{d-1}}|\vec{x}_{t_d}),
\end{align}
for $T$ latents $\vec{x}_{1:T}$ with conditional distributions parameterized by $\theta$. 
The joint distribution of data and latents factorizes into a Markov Chain with parameters
\begin{equation}
    p_{\theta}(\vec{x}_{t_{d-1}}|x_{t_d})=\mathcal{N}(\mu_\theta(\vec{x}_{t_d},t_d), \;\Sigma_\theta(\vec{x}_{t_d}, t_d))), \quad \vec{x}_T~\sim \mathcal{N}(\mathbf{0}, \mat{I})
\end{equation}
which is called the \emph{reverse} process. The posterior $q(\vec{x}_{1:T}|\vec{x}_0)$, called the \emph{forward} process, typically takes the form of a Markov Chain with progressively increasing Gaussian noise parameterized by variance schedule $\beta(t_d)$:
\begin{equation}
    q(\vec{x}_{1:T}|\vec{x}_0)=\prod_{t_d=1}^Tq(\vec{x}_{t_d}|\vec{x}_{t_d-1}), \quad q(\vec{x}_{t_d}|\vec{x}_{t_{d-1}}) = \mathcal{N}(\sqrt{1-\beta(t_d)}\vec{x}_{t_{d-1}}, \;\beta(t_d)\mat{I}))
\end{equation}
where $\beta$ can be either learned or fixed as hyperparameter. The parameters $\theta$ of the reverse process are found by minimizing the variational upper-bound on the negative log-likelihood of the data:
\begin{equation}
    \mathbb{E}_{q}\bigg[-\log p(\vec{x}_T)-\sum_{{t_d}=1}^T\log\frac{p_\theta(\vec{x}_{{t_d}-1}|\vec{x}_{t_d})}{q(\vec{x}_{t_d}|\vec{x}_{{t_d}-1})}\bigg]
    \label{eq:variational_nll}
\end{equation}
Later works, such as Denoising Diffusion Probabilistic Models~\citep[DDPM,][]{ho2020denoising} make specific assumptions regarding the form of $p_\theta$, leading to the following simplified loss with modified variance scale $\bar{\alpha}(t_d)=\prod_{s=1}^{t_d}(1-\beta(s))$:
\begin{equation*}
    \ell_\text{Diffusion}=\mathbb{E}_{\vec{x}_0, t_d, \epsilon}\bigg[|| \epsilon-\epsilon_\theta(\sqrt{\bar{\alpha}(t_d)}\vec{x}_0+\sqrt{1-\bar{\alpha}(t_d)}\epsilon,t_d)||_2^2\bigg],
\end{equation*}
\begin{equation}\vec{x}_0\sim q(\vec{x}_0), \quad t_d\sim \text{Uniform}(1, T),\quad \epsilon\sim \mathcal{N}(\textbf{0}, \mat{I})
    \label{eq:diffusion_loss}
\end{equation}
by training a denoising network $\epsilon_\theta$ to predict  noise $\epsilon$ from a corrupted version of $\vec{x}_0$ at timestep $t_d$. Samples from $p(\vec{x}_0)$ can be generated by following the reverse process:
\begin{equation}
    \vec{x}_{t_d-1} = \frac{1}{\sqrt{\alpha(t_d)}}\bigg(\vec{x}_{t_d}-\frac{1-\alpha(t_d)}{\sqrt{1-\bar{\alpha}}} \epsilon_\theta(\vec{x}_{t_d}, t_d)\bigg) + \sigma_{t_d}\vec{z}, \quad \vec{x}_T\sim \mathcal{N}(\mathbf{0}, \mat{I}), \vec{z}\sim \mathcal{N}(\mathbf{0}, \mat{I}).
\end{equation}

\section{Methodology}

Through the lens of~\cref{eq:occupancy_q_value}, the value function can be decomposed into three components: (1) occupancy measure $\rho^\pi(s)$, dependent on \textbf{states} and \textbf{policy}, (2) reward model $r(s,a)$ dependent on \textbf{states} and \textbf{actions} and (3) policy representation $\phi(\pi)$, dependent on the \textbf{policy}. Equipped with these components, we could estimate the value of any given policy in a zero-shot manner. However, two major issues arise:
\begin{itemize}
    \item For offline\footnote{Online training can use implicit conditioning by re-collecting data with the current policy $\pi$} training, $\rho^\pi$ has to be \emph{explicitly} conditioned on the target policy, via the policy representation $\phi(\pi)$.
    \item Maximizing $Q(s,a,\phi(\pi))$ directly as opposed to indirectly via $r(s,a)+\gamma \mathbb{E}[V(s',\phi(\pi))]$ is too costly due to the large size of diffusion denoising networks.
\end{itemize}

\begin{figure}[h!]
    \centering
    \includegraphics[width=\linewidth]{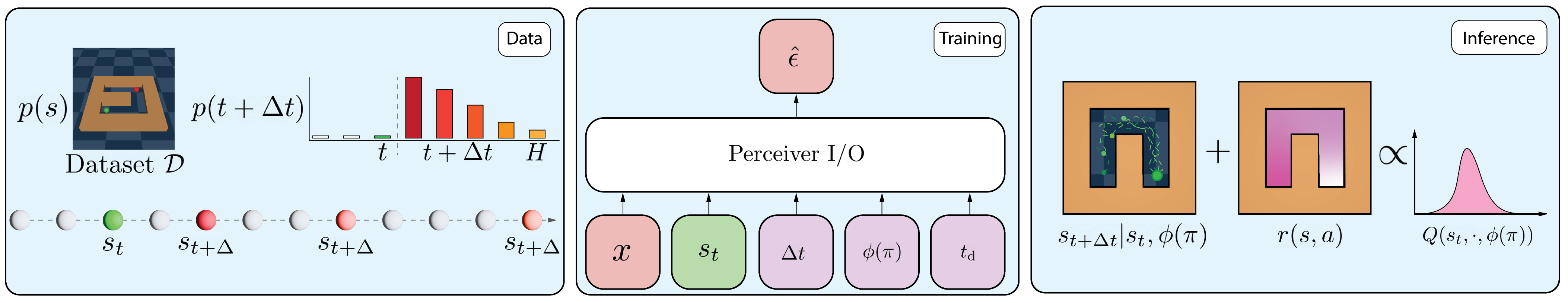}
    \caption{The three crucial components of \shortmethod{}: \textit{(left)} construct tuples $(s_t,s_{t+1},s_{t+\Delta t})$ for training the diffusion model; \textit{(middle)} architecture of the diffusion model, which takes in future noisy state $x$, current state $s_t$, time offset $\Delta t$, policy embedding $\phi(\pi)$ and diffusion timestep $t_d$ and processes them using the Perceiver I/O architecture~\citep{jaegle2021perceiver} to predict the noise;  \textit{(right)} Sampling mechanism based on DPPM~\citep{ho2020denoising} is used with a reward model to estimate the value function}
    \label{fig:model_diagram}
\end{figure}

If both challenges are mitigated, then the value function $V^\pi(s_t)$ can be estimated by first sampling a collection of $n$ states from the learned diffusion model $s_{t+\Delta t,1},..,s_{t+\Delta t,n}\sim \rho^\pi(s_t)$ and then evaluating the reward predictor at those states $\sum_{i=1}^nr(s_{t+\Delta t,i},\pi(s_{t+\Delta t,i}))\propto V^\pi(s_t)$. A similar result can be derived for the state-action value function by training a state-action conditionned diffusion model $\rho^\pi(s_t,a_t)$, from which a policy can be decoded using e.g. the information projection method such as in~\cite{haarnoja2018soft}.

\subsection{Challenge 1: Off-policy evaluation through conditioning}\label{sec:challenge1}
Explicit policy conditioning has (and still remains) a hard task for reinforcement learning settings. Assuming that the policy $\pi$ has a lossless finite-dimensional representation $\phi(\pi)$, passing it to an ideal value function network as $Q(s,a,\phi(\pi))$ could allow for zero-shot policy evaluation. That is, given two policy sets $\Pi_1, \Pi_2\subseteq \Pi$, training $Q(s,a,\phi(\pi))$ on $\{s,a,\phi(\pi)\}, \;\pi\in \Pi_1$ and then swapping out $\phi(\pi)$ for $\phi(\pi')$ where $\pi'\in \Pi_2$ would immediately give the estimate of $Q^{\pi'}$.

We address this issue by studying sufficient statistics of $\pi$. Since the policy is a conditional distribution, it is possible to use a kernel embedding for conditional distributions such as a Reproducing Kernel Hilbert Space~\citep{song2013kernel,mazoure2022low}, albeit it is ill-suited for high-dimensional non-stationary problems. Recent works have studied using the trajectories $\{s_{i},a_i\}_i^n$ as a sufficient statistic for $\pi$ evaluated a \emph{key} states $s_1,..,s_n$~\cite{harb2020policy}. Similarly, we studied two policy representations:
\begin{enumerate}
    \item \textbf{Scalar:} Given a countable policy set $\Pi$ indexed by $i=1,2,..$, we let $\phi(\pi)=i$. One example of such sets is the value improvement path, i.e. the number of training gradient steps performed since initialization.
    \item \textbf{Sequential:} Inspired by~\cite{song2013kernel}, we embed $\pi$ using its rollouts in the environment $\{s_{i},a_i\}_i^n$. In the case where actions are unknown, then the sequence of states can be sufficient, under some mild assumptions\footnote{One such case is MDPs with deterministic dynamics, as it allows to figure out the corresponding action sequence.}, for recovering $\pi$.
\end{enumerate}
Both representations have their own advantages: scalar representations are compact and introduce an ordering into the policy set $\Pi$, while sequential representations can handle cases where no natural ordering is present in $\Pi$ (e.g. learning from offline data).

\subsection{Challenge 2. Maximizing the value with large models}
In domains with continuous actions, the policy is usually decoded using the information projection onto the value function estimate (see~\cite{haarnoja2018soft}) by minimizing
\begin{equation}
    \ell_\text{Policy}(\phi)=\mathbb{E}_{s\sim \cD}\bigg[\text{KL}\bigg(\pi_\phi(\cdot|s)||\frac{e^{Q^{\pi_\text{old}}(s,\cdot)}}{\sum_{a'}e^{Q^{\pi_\text{old}}(s,a')}}\bigg)\bigg]\;.
    \label{eq:policy_loss}
\end{equation}

However, (a) estimating $Q^*(s,a)$ requires estimation of $\rho^*(s,a)$ which cannot be pre-trained on videos (i.e. state sequences) and (b) requires the differentiation of the sampling operator from the $\rho$ network, which, in our work, is parameterized by a large generative model. The same problem arises in both model-free~\citep{haarnoja2018soft} and model-based methods~\citep{hafner2023mastering}, where the networks are sufficiently small that the overhead is minimal. In our work, we circumvent the computational overhead by unrolling one step of Bellman backup
\begin{equation}
    Q^\pi(s_t,a_t) = r(s_t,a_t)+\gamma\mathbb{E}_{s_{t+1}}[V^\pi(s_{t+1})]
\end{equation}
and consequently
\begin{equation}
    \nabla_{a_t} Q^\pi(s_t,a_t) = \nabla_{a_t} r(s_t,a_t),
\end{equation}
allowing to learn $\rho^\pi(s)$ instead of $\rho^\pi(s,a)$ and using it to construct the state value function.

\subsection{Practical algorithm}

As an alternative to classical TD learning, we propose to separately estimate the occupancy measure $\rho$, using a denoising diffusion model and the reward $r$, using a simple regression in symlog space~\citep{hafner2023mastering}. While it is hard to estimate the occupancy measure $\rho$ directly, we instead learn a denoising diffusion probabilistic model $\epsilon_\theta$~\citep[DDPM,][]{ho2020denoising}, which we call the \emph{de-noising network}. Since we know what the true forward process looks like at timestep $t_d$, the de-noising network $\epsilon_\theta: \cS\to [-1,1]$ is trained to predict the input noise.

The high-level idea behind the algorithm is as follows:
\begin{enumerate}
    \item Pre-train a diffusion model on sequences of states $s_1,..,s_H$ and, optionally, policy embeddings $\phi(\pi)$. This step can be performed on large amounts of demonstration videos without the need of any action nor reward labels. The policy embedding $\phi(\pi)$ can be chosen to be any auxiliary information which allows the model to distinguish between policies\footnote{In MDPs with deterministic transitions, one such example is $s_1,..,s_t$.}. This step yields $\rho(s_t;\phi(\pi))$.
    \item Using labeled samples, train a reward predictor $r(s,a)$. This reward predictor will be used as importance weight to score each state-action pair generated by the diffusion model.
    \item Sample a state from $\rho(\cdot,\phi(\pi))$ and score it using $r(s_t, \pi(s_t))$, thus obtaining an estimate proportional to the value function of policy $\pi$ at state $s_t$.
    \item \emph{Optionally:} Maximize the resulting value function estimator using the information projection of $\pi$ onto the polytope of value functions (see~\cref{eq:policy_loss}) and decoding a new policy $\pi'$. If in the online setting, use $\pi'$ to collect new data in the environment and update $\phi(\pi)$ to $\phi(\pi')$.
\end{enumerate}

\begin{figure}[t]
\begin{algorithm}[H]
\SetAlgoLined
\SetKwInOut{Input}{Input}
 \SetKwInOut{Output}{Output}
 \Input{Dataset $\mathcal{D} \sim \mu$, $\epsilon_\theta,r_\psi,\pi_\phi$ networks, number of Monte-Carlo samples $n$}
 \tcc{Normalize states from $\cD$ to lie in $[-1,1]$ interval}
 $\cD[s]\leftarrow \frac{\cD[s]-\min\cD[s]}{\max\cD[s]-\min\cD[s]}$  \;
 \For{epoch $j=1,2,..,J$}{
 \For{minibatch $\cB\sim \mathcal{D}$}{
    \tcc{Update diffusion model $\rho_\theta$ using~\cref{eq:diffusion_loss}}
    Update $\epsilon_\theta$ using $\nabla_{\theta}\ell_\text{Diffusion}(\theta^{(j)})$  over $s_{t+\Delta t}$\;
    \tcc{Update the reward estimator}
    Update $r_\psi$ using $\nabla_\psi \mathbb{E}_{s,a}\bigg[||r_\psi(s,a)-r(s,a)||^2_2\bigg]$ \;
    \tcc{Estimate V}
    $V(s_{t+1})\leftarrow \frac{1-\gamma^{H-t-1}}{1-\gamma}\sum_{i=1}^nr(s_{t+1+\Delta t,i},\pi_\phi(s_{t+1+\Delta t,i}))$, $s_{t+1+\Delta t}\overset{\text{DDPM}}{\sim} \rho_\theta(s_{t+1})$\;
    \tcc{Estimate Q}
    $Q(s_t,a_t)\leftarrow r(s_t,a_t)+\gamma V(s_{t+1})$ \;
    \tcc{Decode policy from Q-function using~\cref{eq:policy_loss}}
    Update $\pi_\phi$ using $\nabla_{\phi}\ell_\text{Policy}(\phi)$ and $Q(s_t,a_t)$ \;
 }
 }
 \caption{\longmethod{} (\shortmethod)}
 \label{alg:offline_algo}
\end{algorithm}
\end{figure}

Algorithm~\ref{alg:offline_algo} describes the exact training mechanism, which first learns a diffusion model and maximizes its reward-weighted expectation to learn a policy suitable for control. Note that the method is suitable for both online and offline reinforcement learning tasks, albeit conditioning on the policy representation $\phi(\pi)$ has to be done explicitly in the case of offline RL.

\shortmethod{} can also be shown to learn an occupancy measure which corresponds to the normalized successor features~\citep{dayan1993improving} that allows sampling future states through the reverse diffusion process.





\section{Experiments}
\subsection{Mountain Car}
\begin{figure}[htbp]
  \centering

  \begin{subfigure}[b]{0.32\textwidth}
    \includegraphics[width=\textwidth]{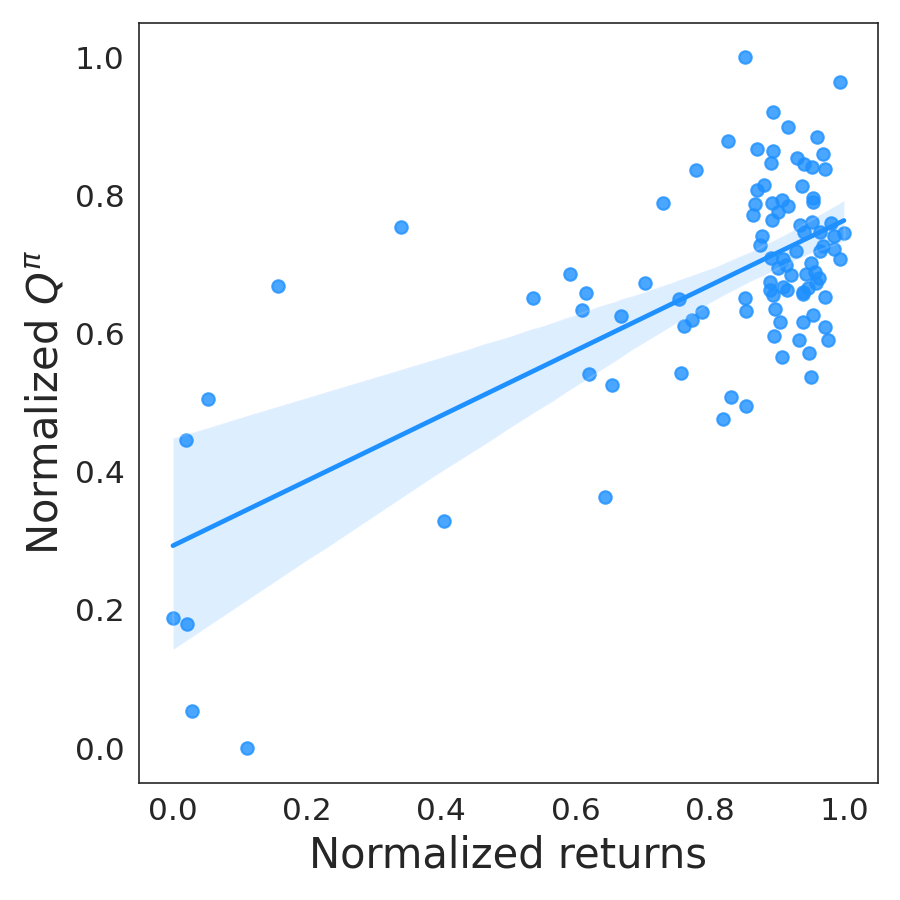}
    \label{fig:returns_vs_Q}
  \end{subfigure}
  \hfill
  \begin{subfigure}[b]{0.32\textwidth}
    \includegraphics[width=\textwidth]{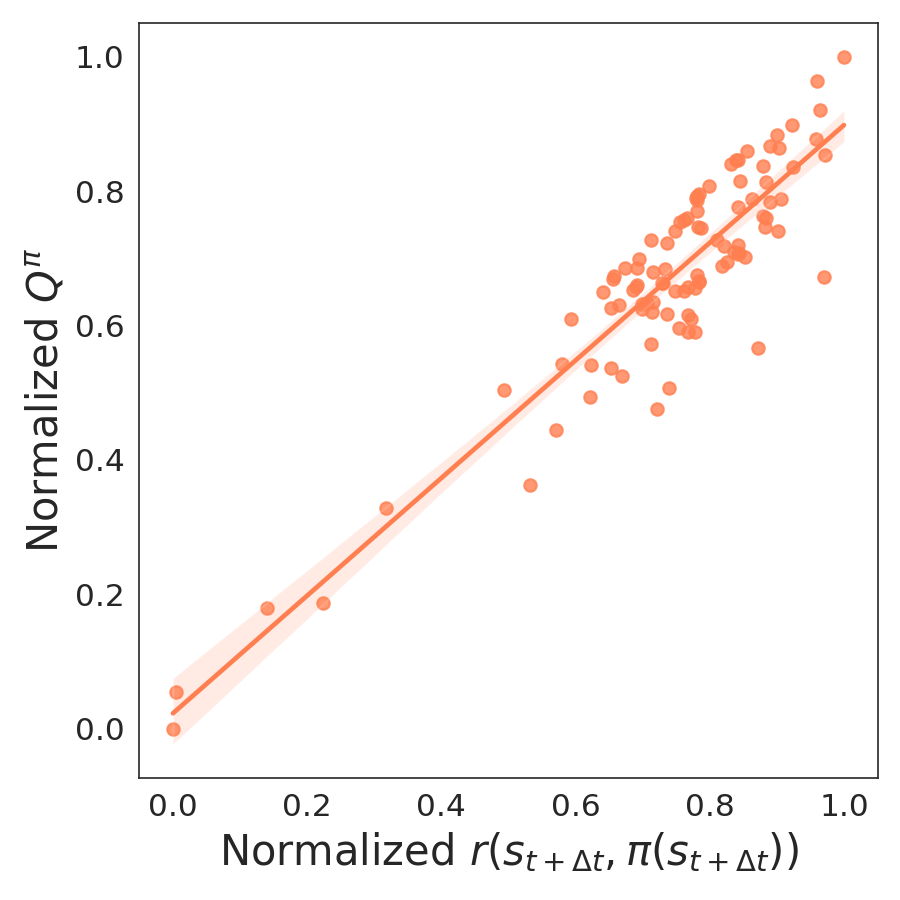}
    \label{fig:rG_vs_Q}
  \end{subfigure}
    \hfill
  \begin{subfigure}[b]{0.32\textwidth}
    \includegraphics[width=\textwidth]{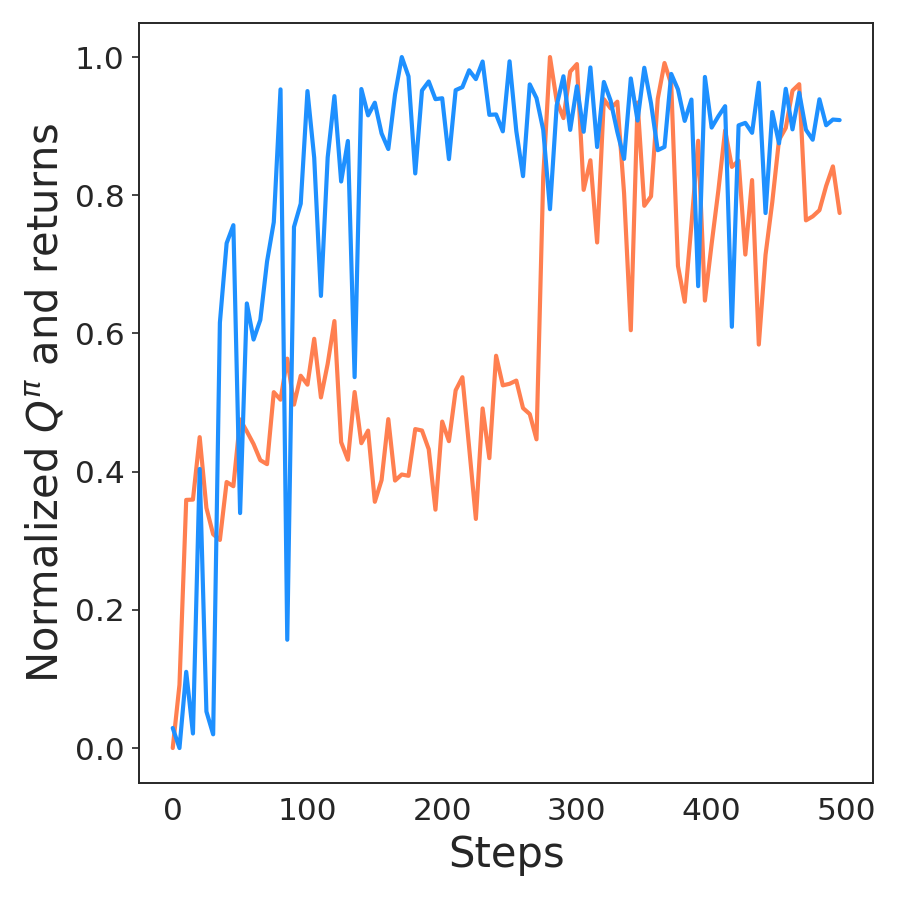}
    \label{fig:mountaincar_curves}
  \end{subfigure}
  \caption{\textit{(Left)} Pairwise plot of normalized returns versus the value function estimated by~\shortmethod{}, \textit{(Middle)} Pairwise plot of normalized value function versus normalized reward at future state and (Right) normalized value function and normalized environment returns versus training gradient steps.}
  \label{fig:mountain_car_regplot}
\end{figure}

Before studying the behavior of \shortmethod{} on robotic tasks, we conduct experiments on the continuous Mountain Car problem, a simple domain for analysing sequential decision making methods. We trained \shortmethod{} for 500 gradient steps until convergence, and computed correlations between the true environment returns, the value function estimator based on the diffusion model, as well as the reward prediction at states sampled from $\rho^\pi$.~\cref{fig:mountain_car_regplot} shows that all three quantities exhibit strong positive correlation, even though the value function estimator is not learned using temporal difference learning.

\subsection{Maze 2d}
\begin{figure}[h!]
    \centering
    \includegraphics[width=\linewidth]{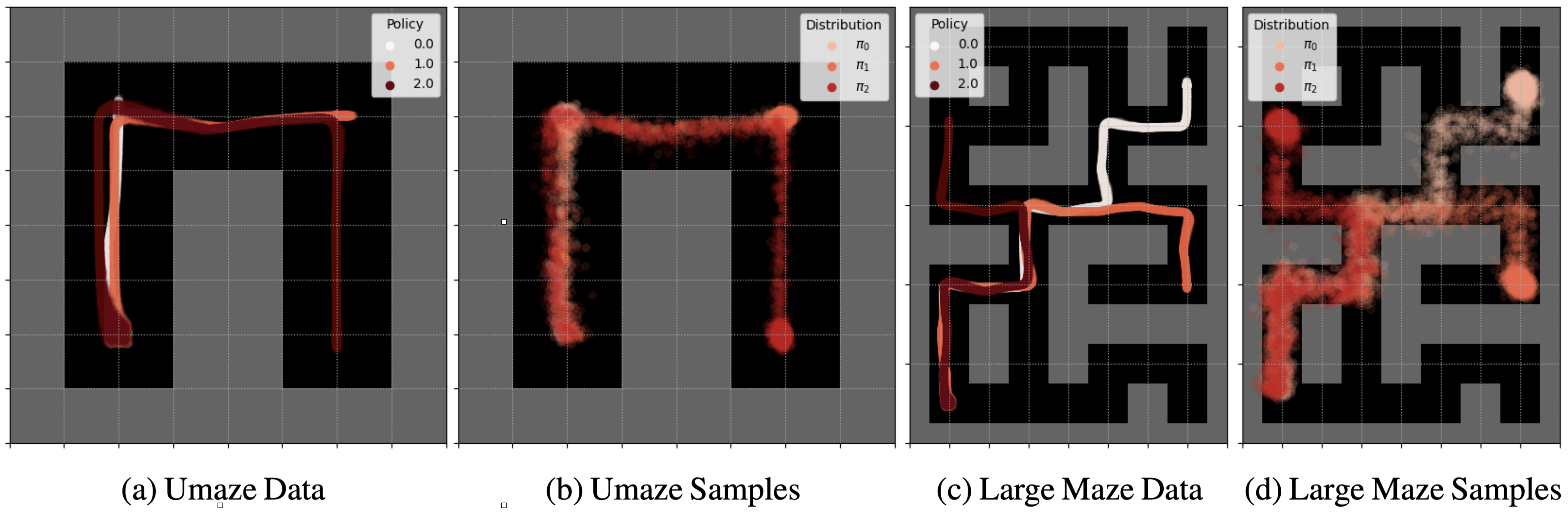}
    \caption{\textit{(a, c)} Ground truth data distribution for the u-maze and large maze from the Maze 2d environment. \textit{(b, d)} Conditional distribution of future states $s_{t+\Delta t}|s_0,\phi(\pi_i)$ given the starting state in the bottom left corner and the policy index. The diffusion model correctly identifies and separates the three state distributions in both mazes.}
    \label{fig:policy_conditionning_maze2d}
\end{figure}

We examine the qualitative behavior of the diffusion model of \shortmethod{} on a simple locomotion task inside mazes of various shapes, as introduced in the D4RL offline suite~\citep{fu2020d4rl}.  In these experiments, the agent starts in the lower left of the maze and uses a waypoint planner with three separate goals to collect data in the environment (see \cref{fig:policy_conditionning_maze2d}(a) and \cref{fig:policy_conditionning_maze2d}(c) for the samples of the collected data).  The diffusion model of \shortmethod{} is trained on the data from the three data-collecting policies, using the scalar policy conditioning described in \cref{sec:challenge1}.

\begin{figure}[h!]
    \centering
    \includegraphics[width=\linewidth]{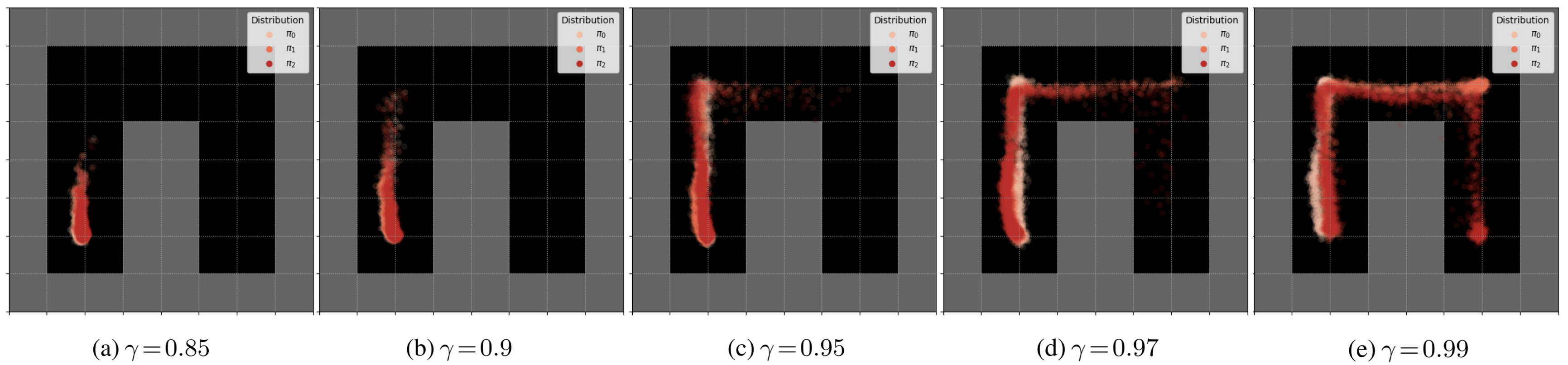}
    \caption{Samples from the learned diffusion model with increasing values of discount factor $\gamma$, with a starting state in the lower left of the maze. As $\gamma$ increases, the model generates samples further along the trajectory leading to the furthest point of the maze.  Ground truth data shown in \cref{fig:policy_conditionning_maze2d}(a) }
    \label{fig:trajectory_samples_maze2d}
\end{figure}

\cref{fig:policy_conditionning_maze2d} shows full trajectories sampled by conditioning the diffusion model on the start state in the lower left, the policy index, and a time offset.  \cref{fig:trajectory_samples_maze2d} shows sampled trajectories as the discount factor $\gamma$ increases, leading to sampling larger time offsets.  

The results show the ability of the diffusion model to represent long-horizon data faithfully, and highlight some benefits of the approach.  \shortmethod{} can sample trajectories without the need to evaluate a policy or specify intermediate actions.  Because \shortmethod{} samples each time offset independently, there is also no concern of compounding model error as the horizon increases. Additionally, the cost of predicting $s_{t+k}$ from $s_t$ is $\cO(1)$ for \shortmethod{}, while it is $\mathcal{O}(k)$ for classical autoregressive models.

\subsection{PyBullet}
Our final set of experiments consists in ablations performed on offline data collected from classical PyBullet environments\footnote{Data is taken from \url{https://github.com/takuseno/d4rl-pybullet/tree/master}}, as oppposed to D4RL, which has faced criticism due to poor sim2real transfer capabilities~\citep{korber2021comparing}. We compare \shortmethod{} to behavior cloning and Conservative Q-learning~\citep{kumar2020conservative}, two strong offline RL baselines. We also plot the average normalized returns of each dataset to facilitate the comparison over 5 random seeds. Medium dataset contains data collected by medium-level policy and mixed contains data from SAC~\citep{haarnoja2018soft} training.

\begin{figure}[h!]
    \centering
    \includegraphics[width=\linewidth]{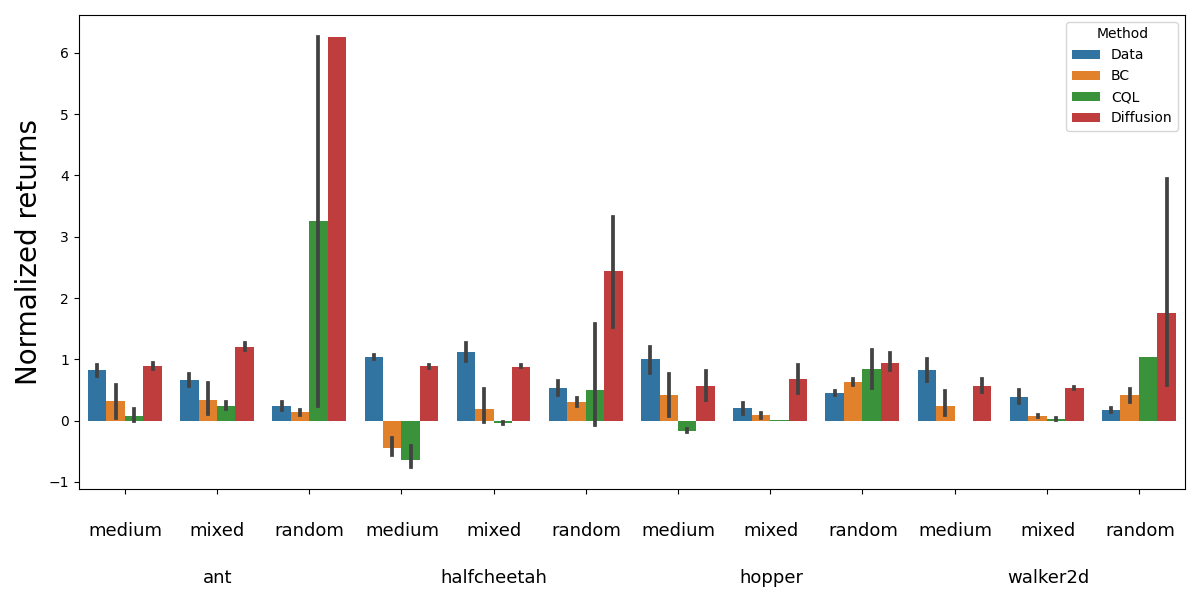}
    \caption{Normalized returns obtained by \shortmethod{}, behavior cloning, CQL on 4 challenging robotic tasks from the PyBullet offline suite, together with average returns in each dataset (\textit{Data} in the plot).}
    \label{fig:pybullet_hist}
\end{figure}

\cref{fig:pybullet_hist} highlights the ability of \shortmethod{} to match and sometimes outperform the performance of classical offline RL algorithms, especially on data of lower quality, e.g. coming from a random policy. In domains where online rollouts can be prohibitively expensive, the ability to learn from low-quality, incomplete offline demonstrations is a strength of \shortmethod{}. This benchmark also demonstrates the shortcomings of scalar policy representation, which is unknown for a given offline dataset, and also doesn't scale well when the number of policies is large (e.g. number of gradient steps $j$ in logging policy $\mu^{(j)}$). For this reason, we opted for a sequential policy representation.

\section{Related works}

\paragraph{Offline pre-training for reinforcement learning} Multiple approaches have tried to alleviate the heavy cost of training agents \emph{tabula rasa} by pre-training some parts of the system offline. For example, inverse dynamics models which predict the action leading from the current state to the next state have seen success in complex domains such as Atari~\citep{schwarzer2021pretraining}, as well as Minecraft~\citep{baker2022video, fan2022minedojo}. Return-conditioned sequence models have also seen a rise in popularity, specifically due to their ability to learn performance-action-state correlations over long horizons~\citep{lee2022multi}. 


\paragraph{Unsupervised reinforcement learning}
Using temporal difference~\citep{sutton2018reinforcement} for policy iteration or evaluation requires all data tuples to contain state, action and reward information. However, in some real-world scenarios, the reward might only be available for a small subset of data (e.g. problems with delayed feedback~\citep{howson2021delayed}). In this case, it is possible to decompose the value function into a reward-dependent and dynamics components, as was first suggested in the successor representation framework~\citep{dayan1993improving,barreto2016successor}. More recent approaches~\citep{janner2020generative,eysenbach2020c,eysenbach2022contrastive,mazoure2022contrastive} use a density model to learn the occupancy measure over future states for each state-action pair in the dataset. However, learning an explicit multi-step model such as~\citep{janner2020generative} can be unstable due to the bootstrapping term in the temporal difference loss, and these approaches still require large amounts of reward and action labels. While our proposed method is a hybrid between model-free and model-based learning, it avoids the computational overhead incurred by classical world models such as Dreamer~\citep{hafner2023mastering} by introducing constant-time rollouts. The main issue with infinite-horizon models is the implicit dependence of the model on the policy, which imposes an upper-bound on the magnitude of the policy improvement step achievable in the offline case. Our work solves this issue by adding an explicit policy conditioning mechanism, which allows to generate future states from unseen policy embeddings.

\paragraph{Diffusion models}
Learning a conditional probability distribution over a highly complex space can be a challenging task, which is why it is often easier to instead approximate it using a density ratio specified by an inner product in a much lower-dimensional latent space. To learn an occupancy measure over future states without passing via the temporal difference route, one can use denoising diffusion models to approximate the corresponding future state density under a given policy. Diffusion has previously been used in the static unsupervised setting such as image generation~\citep{ho2020denoising} and text-to-image generation~\citep{rombach2022high}. Diffusion models have also been used to model trajectory data for planning in small-dimensional environments~\citep{janner2022planning}. However, no work so far has managed to efficiently predict infinite-horizon rollouts.

\section{Discussion}
In this work, we introduced a simple model-free algorithm for learning reward-maximizing policies, which can be efficiently used to solve complex robotic tasks. \longmethod{} (\shortmethod) avoids the pitfalls of both temporal difference learning and autoregressive model-based methods by pre-training an infinite-horizon transition model from state sequences using a diffusion model. This model does not require any action nor reward information, and can then be used to construct the state-action value function, from which one can decode the optimal action. \shortmethod{} fully leverages the power of diffusion models to generate states far ahead into the future without intermediate predictions. Our experiments demonstrate that \shortmethod{} matches and sometimes outperforms strong offline RL baselines on realistic robotic tasks based on PyBullet, and opens an entire new direction of research.

\section{Limitations}
The main limitation of our method is that it operates directly on observations instead of latent state embeddings, which requires tuning the noise schedule for each set of tasks, instead of using a unified noise schedule similarly to latent diffusion models~\citep{Rombach_2022_CVPR}. Another limitation is the need to explicitly condition the rollouts from the diffusion model on the policy, something that single-step models avoid. Finally, online learning introduces the challenge of capturing the non-stationarity of the environment using the generative model $\rho$, which, in itself, is a hard task.


\begin{thebibliography}{37}
\providecommand{\natexlab}[1]{#1}
\providecommand{\url}[1]{\texttt{#1}}
\expandafter\ifx\csname urlstyle\endcsname\relax
  \providecommand{\doi}[1]{doi: #1}\else
  \providecommand{\doi}{doi: \begingroup \urlstyle{rm}\Url}\fi

\bibitem[Chowdhery et~al.(2022)Chowdhery, Narang, Devlin, Bosma, Mishra,
  Roberts, Barham, Chung, Sutton, Gehrmann, et~al.]{chowdhery2022palm}
A.~Chowdhery, S.~Narang, J.~Devlin, M.~Bosma, G.~Mishra, A.~Roberts, P.~Barham,
  H.~W. Chung, C.~Sutton, S.~Gehrmann, et~al.
\newblock Palm: Scaling language modeling with pathways.
\newblock \emph{arXiv preprint arXiv:2204.02311}, 2022.

\bibitem[Touvron et~al.(2023)Touvron, Lavril, Izacard, Martinet, Lachaux,
  Lacroix, Rozi{\`e}re, Goyal, Hambro, Azhar, et~al.]{touvron2023llama}
H.~Touvron, T.~Lavril, G.~Izacard, X.~Martinet, M.-A. Lachaux, T.~Lacroix,
  B.~Rozi{\`e}re, N.~Goyal, E.~Hambro, F.~Azhar, et~al.
\newblock Llama: Open and efficient foundation language models.
\newblock \emph{arXiv preprint arXiv:2302.13971}, 2023.

\bibitem[Kaplan et~al.(2020)Kaplan, McCandlish, Henighan, Brown, Chess, Child,
  Gray, Radford, Wu, and Amodei]{kaplan2020scaling}
J.~Kaplan, S.~McCandlish, T.~Henighan, T.~B. Brown, B.~Chess, R.~Child,
  S.~Gray, A.~Radford, J.~Wu, and D.~Amodei.
\newblock Scaling laws for neural language models.
\newblock \emph{arXiv preprint arXiv:2001.08361}, 2020.

\bibitem[Ouyang et~al.(2022)Ouyang, Wu, Jiang, Almeida, Wainwright, Mishkin,
  Zhang, Agarwal, Slama, Ray, et~al.]{ouyang2022training}
L.~Ouyang, J.~Wu, X.~Jiang, D.~Almeida, C.~Wainwright, P.~Mishkin, C.~Zhang,
  S.~Agarwal, K.~Slama, A.~Ray, et~al.
\newblock Training language models to follow instructions with human feedback.
\newblock \emph{Advances in Neural Information Processing Systems},
  35:\penalty0 27730--27744, 2022.

\bibitem[Brohan et~al.(2023)Brohan, Chebotar, Finn, Hausman, Herzog, Ho, Ibarz,
  Irpan, Jang, Julian, et~al.]{brohan2023can}
A.~Brohan, Y.~Chebotar, C.~Finn, K.~Hausman, A.~Herzog, D.~Ho, J.~Ibarz,
  A.~Irpan, E.~Jang, R.~Julian, et~al.
\newblock Do as i can, not as i say: Grounding language in robotic affordances.
\newblock In \emph{Conference on Robot Learning}, pages 287--318. PMLR, 2023.

\bibitem[Stone et~al.(2023)Stone, Xiao, Lu, Gopalakrishnan, Lee, Vuong,
  Wohlhart, Zitkovich, Xia, Finn, et~al.]{stone2023open}
A.~Stone, T.~Xiao, Y.~Lu, K.~Gopalakrishnan, K.-H. Lee, Q.~Vuong, P.~Wohlhart,
  B.~Zitkovich, F.~Xia, C.~Finn, et~al.
\newblock Open-world object manipulation using pre-trained vision-language
  models.
\newblock \emph{arXiv preprint arXiv:2303.00905}, 2023.

\bibitem[Driess et~al.(2023)Driess, Xia, Sajjadi, Lynch, Chowdhery, Ichter,
  Wahid, Tompson, Vuong, Yu, et~al.]{driess2023palm}
D.~Driess, F.~Xia, M.~S. Sajjadi, C.~Lynch, A.~Chowdhery, B.~Ichter, A.~Wahid,
  J.~Tompson, Q.~Vuong, T.~Yu, et~al.
\newblock Palm-e: An embodied multimodal language model.
\newblock \emph{arXiv preprint arXiv:2303.03378}, 2023.

\bibitem[Baker et~al.(2022)Baker, Akkaya, Zhokov, Huizinga, Tang, Ecoffet,
  Houghton, Sampedro, and Clune]{baker2022video}
B.~Baker, I.~Akkaya, P.~Zhokov, J.~Huizinga, J.~Tang, A.~Ecoffet, B.~Houghton,
  R.~Sampedro, and J.~Clune.
\newblock Video pretraining (vpt): Learning to act by watching unlabeled online
  videos.
\newblock \emph{Advances in Neural Information Processing Systems},
  35:\penalty0 24639--24654, 2022.

\bibitem[Fan et~al.(2022)Fan, Wang, Jiang, Mandlekar, Yang, Zhu, Tang, Huang,
  Zhu, and Anandkumar]{fan2022minedojo}
L.~Fan, G.~Wang, Y.~Jiang, A.~Mandlekar, Y.~Yang, H.~Zhu, A.~Tang, D.-A. Huang,
  Y.~Zhu, and A.~Anandkumar.
\newblock Minedojo: Building open-ended embodied agents with internet-scale
  knowledge.
\newblock \emph{arXiv preprint arXiv:2206.08853}, 2022.

\bibitem[Yu et~al.(2020)Yu, Thomas, Yu, Ermon, Zou, Levine, Finn, and
  Ma]{yu2020mopo}
T.~Yu, G.~Thomas, L.~Yu, S.~Ermon, J.~Y. Zou, S.~Levine, C.~Finn, and T.~Ma.
\newblock Mopo: Model-based offline policy optimization.
\newblock \emph{Advances in Neural Information Processing Systems},
  33:\penalty0 14129--14142, 2020.

\bibitem[Argenson and Dulac-Arnold(2020)]{argenson2020model}
A.~Argenson and G.~Dulac-Arnold.
\newblock Model-based offline planning.
\newblock \emph{arXiv preprint arXiv:2008.05556}, 2020.

\bibitem[Kidambi et~al.(2020)Kidambi, Rajeswaran, Netrapalli, and
  Joachims]{kidambi2020morel}
R.~Kidambi, A.~Rajeswaran, P.~Netrapalli, and T.~Joachims.
\newblock Morel: Model-based offline reinforcement learning.
\newblock \emph{arXiv preprint arXiv:2005.05951}, 2020.

\bibitem[Yu et~al.(2021)Yu, Kumar, Rafailov, Rajeswaran, Levine, and
  Finn]{yu2021combo}
T.~Yu, A.~Kumar, R.~Rafailov, A.~Rajeswaran, S.~Levine, and C.~Finn.
\newblock Combo: Conservative offline model-based policy optimization.
\newblock \emph{Advances in neural information processing systems},
  34:\penalty0 28954--28967, 2021.

\bibitem[Dayan(1993)]{dayan1993improving}
P.~Dayan.
\newblock Improving generalization for temporal difference learning: The
  successor representation.
\newblock \emph{Neural Computation}, 5\penalty0 (4):\penalty0 613--624, 1993.

\bibitem[Janner et~al.(2020)Janner, Mordatch, and Levine]{janner2020generative}
M.~Janner, I.~Mordatch, and S.~Levine.
\newblock Generative temporal difference learning for infinite-horizon
  prediction.
\newblock \emph{arXiv preprint arXiv:2010.14496}, 2020.

\bibitem[Barreto et~al.(2016)Barreto, Dabney, Munos, Hunt, Schaul, Van~Hasselt,
  and Silver]{barreto2016successor}
A.~Barreto, W.~Dabney, R.~Munos, J.~J. Hunt, T.~Schaul, H.~Van~Hasselt, and
  D.~Silver.
\newblock Successor features for transfer in reinforcement learning.
\newblock \emph{arXiv preprint arXiv:1606.05312}, 2016.

\bibitem[Sohl-Dickstein et~al.(2015)Sohl-Dickstein, Weiss, Maheswaranathan, and
  Ganguli]{sohl2015deep}
J.~Sohl-Dickstein, E.~Weiss, N.~Maheswaranathan, and S.~Ganguli.
\newblock Deep unsupervised learning using nonequilibrium thermodynamics.
\newblock In \emph{International Conference on Machine Learning}, pages
  2256--2265. PMLR, 2015.

\bibitem[Ho et~al.(2020)Ho, Jain, and Abbeel]{ho2020denoising}
J.~Ho, A.~Jain, and P.~Abbeel.
\newblock Denoising diffusion probabilistic models.
\newblock \emph{Advances in Neural Information Processing Systems},
  33:\penalty0 6840--6851, 2020.

\bibitem[Jaegle et~al.(2021)Jaegle, Borgeaud, Alayrac, Doersch, Ionescu, Ding,
  Koppula, Zoran, Brock, Shelhamer, et~al.]{jaegle2021perceiver}
A.~Jaegle, S.~Borgeaud, J.-B. Alayrac, C.~Doersch, C.~Ionescu, D.~Ding,
  S.~Koppula, D.~Zoran, A.~Brock, E.~Shelhamer, et~al.
\newblock Perceiver io: A general architecture for structured inputs \&
  outputs.
\newblock \emph{arXiv preprint arXiv:2107.14795}, 2021.

\bibitem[Haarnoja et~al.(2018)Haarnoja, Zhou, Abbeel, and
  Levine]{haarnoja2018soft}
T.~Haarnoja, A.~Zhou, P.~Abbeel, and S.~Levine.
\newblock Soft actor-critic: Off-policy maximum entropy deep reinforcement
  learning with a stochastic actor.
\newblock In \emph{International conference on machine learning}, pages
  1861--1870. PMLR, 2018.

\bibitem[Song et~al.(2013)Song, Fukumizu, and Gretton]{song2013kernel}
L.~Song, K.~Fukumizu, and A.~Gretton.
\newblock Kernel embeddings of conditional distributions: A unified kernel
  framework for nonparametric inference in graphical models.
\newblock \emph{IEEE Signal Processing Magazine}, 30\penalty0 (4):\penalty0
  98--111, 2013.

\bibitem[Mazoure et~al.(2022)Mazoure, Doan, Li, Makarenkov, Pineau, Precup, and
  Rabusseau]{mazoure2022low}
B.~Mazoure, T.~Doan, T.~Li, V.~Makarenkov, J.~Pineau, D.~Precup, and
  G.~Rabusseau.
\newblock Low-rank representation of reinforcement learning policies.
\newblock \emph{Journal of Artificial Intelligence Research}, 75:\penalty0
  597--636, 2022.

\bibitem[Harb et~al.(2020)Harb, Schaul, Precup, and Bacon]{harb2020policy}
J.~Harb, T.~Schaul, D.~Precup, and P.-L. Bacon.
\newblock Policy evaluation networks.
\newblock \emph{arXiv preprint arXiv:2002.11833}, 2020.

\bibitem[Hafner et~al.(2023)Hafner, Pasukonis, Ba, and
  Lillicrap]{hafner2023mastering}
D.~Hafner, J.~Pasukonis, J.~Ba, and T.~Lillicrap.
\newblock Mastering diverse domains through world models.
\newblock \emph{arXiv preprint arXiv:2301.04104}, 2023.

\bibitem[Fu et~al.(2020)Fu, Kumar, Nachum, Tucker, and Levine]{fu2020d4rl}
J.~Fu, A.~Kumar, O.~Nachum, G.~Tucker, and S.~Levine.
\newblock D4rl: Datasets for deep data-driven reinforcement learning, 2020.

\bibitem[K{\"o}rber et~al.(2021)K{\"o}rber, Lange, Rediske, Steinmann, and
  Gl{\"u}ck]{korber2021comparing}
M.~K{\"o}rber, J.~Lange, S.~Rediske, S.~Steinmann, and R.~Gl{\"u}ck.
\newblock Comparing popular simulation environments in the scope of robotics
  and reinforcement learning.
\newblock \emph{arXiv preprint arXiv:2103.04616}, 2021.

\bibitem[Kumar et~al.(2020)Kumar, Zhou, Tucker, and
  Levine]{kumar2020conservative}
A.~Kumar, A.~Zhou, G.~Tucker, and S.~Levine.
\newblock Conservative q-learning for offline reinforcement learning.
\newblock \emph{arXiv preprint arXiv:2006.04779}, 2020.

\bibitem[Schwarzer et~al.(2021)Schwarzer, Rajkumar, Noukhovitch, Anand,
  Charlin, Hjelm, Bachman, and Courville]{schwarzer2021pretraining}
M.~Schwarzer, N.~Rajkumar, M.~Noukhovitch, A.~Anand, L.~Charlin, D.~Hjelm,
  P.~Bachman, and A.~Courville.
\newblock Pretraining representations for data-efficient reinforcement
  learning.
\newblock \emph{arXiv preprint arXiv:2106.04799}, 2021.

\bibitem[Lee et~al.(2022)Lee, Nachum, Yang, Lee, Freeman, Guadarrama, Fischer,
  Xu, Jang, Michalewski, et~al.]{lee2022multi}
K.-H. Lee, O.~Nachum, M.~S. Yang, L.~Lee, D.~Freeman, S.~Guadarrama,
  I.~Fischer, W.~Xu, E.~Jang, H.~Michalewski, et~al.
\newblock Multi-game decision transformers.
\newblock \emph{Advances in Neural Information Processing Systems},
  35:\penalty0 27921--27936, 2022.

\bibitem[Sutton and Barto(2018)]{sutton2018reinforcement}
R.~S. Sutton and A.~G. Barto.
\newblock \emph{Reinforcement learning: An introduction}.
\newblock MIT press, 2018.

\bibitem[Howson et~al.(2021)Howson, Pike-Burke, and Filippi]{howson2021delayed}
B.~Howson, C.~Pike-Burke, and S.~Filippi.
\newblock Delayed feedback in episodic reinforcement learning.
\newblock \emph{arXiv preprint arXiv:2111.07615}, 2021.

\bibitem[Eysenbach et~al.(2020)Eysenbach, Salakhutdinov, and
  Levine]{eysenbach2020c}
B.~Eysenbach, R.~Salakhutdinov, and S.~Levine.
\newblock C-learning: Learning to achieve goals via recursive classification.
\newblock \emph{arXiv preprint arXiv:2011.08909}, 2020.

\bibitem[Eysenbach et~al.(2022)Eysenbach, Zhang, Salakhutdinov, and
  Levine]{eysenbach2022contrastive}
B.~Eysenbach, T.~Zhang, R.~Salakhutdinov, and S.~Levine.
\newblock Contrastive learning as goal-conditioned reinforcement learning.
\newblock \emph{arXiv preprint arXiv:2206.07568}, 2022.

\bibitem[Mazoure et~al.(2022)Mazoure, Eysenbach, Nachum, and
  Tompson]{mazoure2022contrastive}
B.~Mazoure, B.~Eysenbach, O.~Nachum, and J.~Tompson.
\newblock Contrastive value learning: Implicit models for simple offline rl,
  2022.

\bibitem[Rombach et~al.(2022)Rombach, Blattmann, Lorenz, Esser, and
  Ommer]{rombach2022high}
R.~Rombach, A.~Blattmann, D.~Lorenz, P.~Esser, and B.~Ommer.
\newblock High-resolution image synthesis with latent diffusion models.
\newblock In \emph{Proceedings of the IEEE/CVF Conference on Computer Vision
  and Pattern Recognition}, pages 10684--10695, 2022.

\bibitem[Janner et~al.(2022)Janner, Du, Tenenbaum, and
  Levine]{janner2022planning}
M.~Janner, Y.~Du, J.~B. Tenenbaum, and S.~Levine.
\newblock Planning with diffusion for flexible behavior synthesis.
\newblock \emph{arXiv preprint arXiv:2205.09991}, 2022.

\bibitem[Rombach et~al.(2022)Rombach, Blattmann, Lorenz, Esser, and
  Ommer]{Rombach_2022_CVPR}
R.~Rombach, A.~Blattmann, D.~Lorenz, P.~Esser, and B.~Ommer.
\newblock High-resolution image synthesis with latent diffusion models.
\newblock In \emph{Proceedings of the IEEE/CVF Conference on Computer Vision
  and Pattern Recognition (CVPR)}, pages 10684--10695, June 2022.

\end{thebibliography}

\newpage

\section*{Appendix}
\subsection{Experimental details}
\label{sec:experimental_details}

\paragraph{Model architecture} \shortmethod{} uses a Perceiver I/O model~\citep{jaegle2021perceiver} with $1\times 1$ convolution encodings for states, sinusoidal encoding for diffusion timestep and a linear layer for action embedding. The Perceiver I/O model has positional encodings for all inputs, followed by 8 blocks with 4 cross-attention heads and 4 self-attention heads and latent size 256. The scalar policy representation was encoded using sinusoidal encoding, while the sequential representation was passed through the $1\times 1$ convolution and linear embedding layers and masked-out to handle varying context lengths, before being passed to the Perceiver model.

\begin{table}[h]
    \centering
    \begin{tabular}{l|l}
    \toprule
    Hyperparameter & Value \\
    \midrule
    \midrule
    Learning rate & $3 \times 10^{-4}$ \\
    Batch size & $128$ \\
    Discount factor & 0.99 \\
    Max gradient norm & 100\\
    MLP structure & $256\times 256$ DenseNet MLP\\
    Add LayerNorm in between all layers & Yes\\
    \bottomrule
    \end{tabular}
    \caption{Hyperparameters that are consistent between methods.}
    \label{tab:common_hps}
\end{table}

\begin{table}[h]
    \centering
    \begin{tabular}{l|l}
    \toprule
    Hyperparameter & Value \\
    \midrule
    \midrule
    \shortmethod{} & \\
    \midrule
    Number of future state samples $n$ & 32\\
    BC coefficient & 0.1 (offline) and 0 (online)\\
    \midrule
    CQL & \\
    \midrule
    Regularization coefficient & 1\\
    \bottomrule
    \end{tabular}
    \caption{Hyperparameters that are different between methods.}
    \label{tab:method_hps}
\end{table}

All experiments were run on the equivalent of 2 V100 GPUs with 32 Gb of VRAM and 8 CPUs.

\paragraph{Dataset composition} The Maze2d datasets were constructed based on waypoint planning scripts provided in the D4RL repository, and modifying the target goal locations to lie in each corner of the maze (u-maze), or in randomly chosen pathways (large maze). The PyBullet dataset has a data composition similar to the original D4RL suite, albeit collected in the PyBullet simulator instead of MuJoCo.

\subsection{Additional results} We include three videos of the training of the diffusion model $\rho$ on the large maze dataset shown in~\cref{fig:trajectory_samples_maze2d} for 128, 512 and 1024 diffusion timesteps, in the supplementary material. Note that increasing the number of timesteps leads to faster convergence of the diffusion model samples to the true data distribution.

\end{document}